\documentclass[sigconf,screen,nonacm]{acmart}
\usepackage{amsmath,amsfonts}
\usepackage{graphicx}
\usepackage{textcomp}
\usepackage{xcolor}
\usepackage{hhline}
\usepackage{algorithmicx}
\usepackage{algorithm}
\usepackage{algpseudocode}
\usepackage{caption}
\usepackage{multirow}
\usepackage{fancyvrb}
\usepackage{float}
\usepackage{subfig}
\usepackage{enumitem}
\usepackage{blindtext}
\usepackage{hyperref}
\usepackage[belowskip=-3pt,aboveskip=0pt]{caption}
\makeatletter
\def\maxwidth#1{\ifdim\Gin@nat@width>#1 #1\else\Gin@nat@width\fi}
\makeatother
\setlist[itemize]{topsep=3pt}
\AtBeginDocument{%
  }

\setcopyright{acmlicensed}
\copyrightyear{2018}
\acmYear{2018}
\acmDOI{XXXXXXX.XXXXXXX}

\begin{document}

\title{TryOffAnyone: Tiled Cloth Generation from a Dressed Person}



\author{Ioannis Xarchakos}
\email{ioannis.xarchakos@sabinodb.com}
\additionalaffiliation{University of Toronto}
\affiliation{%
  \institution{SabinoDB}
  \city{Athens}
  \country{Greece}}

\author{Theodoros Koukopoulos}
\email{thodoris.koukopoulos@sabinodb.com}
\affiliation{%
  \institution{SabinoDB}
  \city{Athens}
  \country{Greece}}

\begin{abstract}

The fashion industry is increasingly leveraging computer vision and deep learning technologies to enhance online shopping experiences and operational efficiencies. In this paper, we address the challenge of generating high-fidelity tiled garment images—essential for personalized recommendations, outfit composition, and virtual try-on systems—from photos of garments worn by models. 

Inspired by the success of Latent Diffusion Models (LDMs) in image-to-image translation, we propose a novel approach utilizing a fine-tuned StableDiffusion model. Our method features a streamlined single-stage network design, which integrates garment-specific masks to isolate and process target clothing items effectively. By simplifying the network architecture through selective training of transformer blocks and removing unnecessary cross-attention layers, we significantly reduce computational complexity while achieving state-of-the-art performance on benchmark datasets like VITON-HD. Experimental results demonstrate the effectiveness of our approach in producing high-quality tiled garment images for both full-body and half-body inputs.  Code and model are available at: \href{https://github.com/ixarchakos/try-off-anyone}{https://github.com/ixarchakos/try-off-anyone}

\end{abstract}

\maketitle
\section{Introduction}
\label{sec:intro}

The fashion industry constitutes a critical component of the global economy, generating over \$1.7 trillion annually. Beyond fueling economic growth, it influences cultural trends and societal norms worldwide. Adopting technologies such as artificial intelligence, computer vision, and machine learning is enhancing operational efficiency, streamlining supply chains, and reducing costs in ways that were previously unattainable. These innovations enable precise demand forecasting, improved inventory management, and optimized distribution, addressing long-standing industry challenges such as overproduction and waste. Additionally, computer vision is empowering new user-centric applications, from personalized shopping experiences and virtual try-ons to enhanced product recommendations, which collectively improve customer engagement and drive revenue growth.

\begin{figure}
\centering
\setkeys{Gin}{width=0.38\linewidth}
\subfloat[Tiled cloth image]{\includegraphics{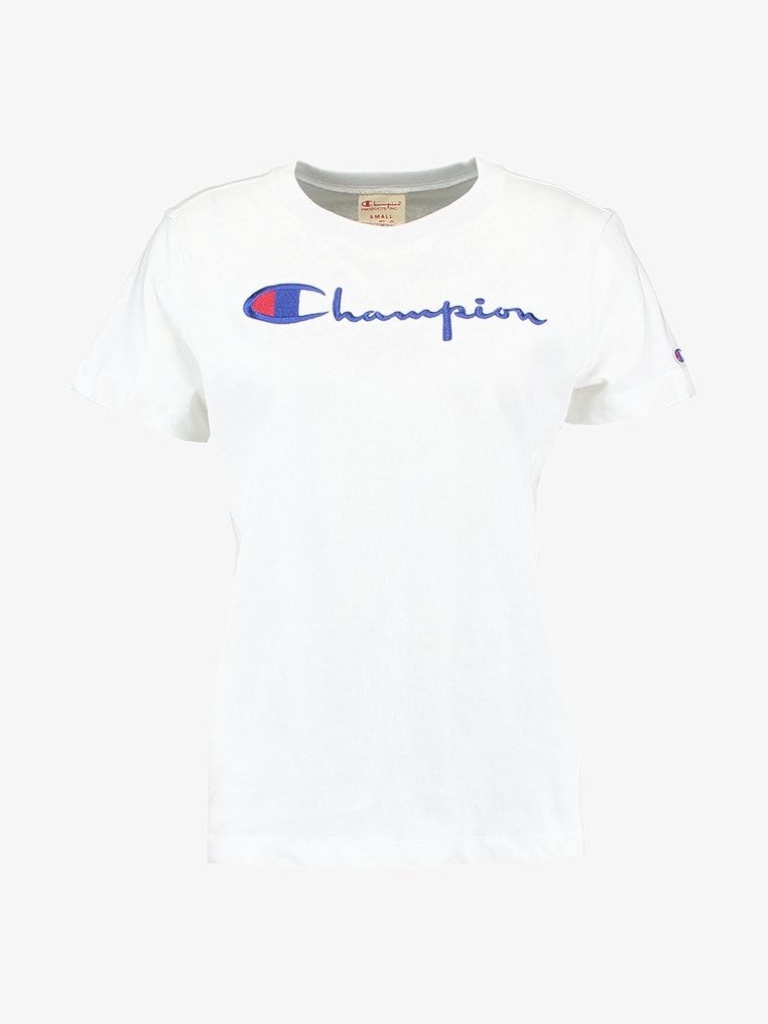}}\hfil
\subfloat[Human worn cloth image]{\includegraphics{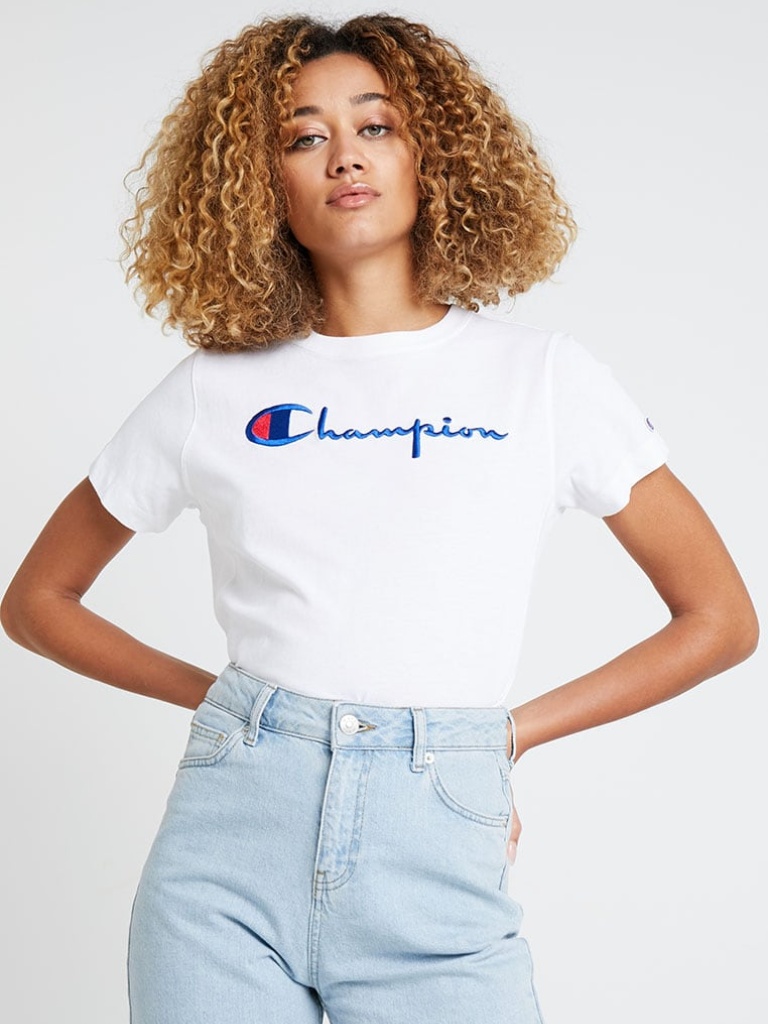}}\hfil
\caption{\centering Different garment image views}
\label{fig:garment}
\end{figure}

Recent advances in Deep Learning (DL) have transformed many applications in computer vision and natural language processing. State-of-the-art algorithms have been introduced for image classification, object detection, semantic segmentation, and, image generation. This rapid progress has significantly impacted the fashion industry, driving the development of applications such as product attribution, image retrieval, outfit recommendation, and virtual try-on systems \cite{ding2023computational, cheng2021fashion}. A core element of these applications is the tiled view (or lay-downs) of garment images (Figure \ref{fig:garment}a), which serves as the primary input format. These tiled views are essential for enhancing the online shopping experience by offering users similar items, outfit recommendations, and even virtual try-ons, thereby creating a more immersive and personalized shopping experience.

While many online shopping platforms showcase their clothing through images of garments worn by models (Figure \ref{fig:garment}b), few include the lay-down view of the garment. This limitation restricts their ability to feature these views on their websites and hinders key applications designed to improve user experience. Acquiring additional tiled images can be costly and labor-intensive, posing a barrier for many retailers. To address this, recent advancements in computer vision have investigated image-to-image translation techniques, particularly using Generative Adversarial Networks (GANs) \cite{zeng2020tilegan} and Latent Diffusion Models \cite{velioglu2024tryoffdiffvirtualtryoffhighfidelitygarment}. These techniques enable the generation of tiled garment images from photos of clothing worn on models, offering a cost-effective solution for creating tiled cloth views or virtual try-offs (VTOFF) \cite{velioglu2024tryoffdiffvirtualtryoffhighfidelitygarment}.

Inspired by the recent success of Latent Diffusion Models (LDMs) in image-to-image translation tasks such as super-resolution, colorization, inpainting, and virtual try-on \cite{lugmayr2022palette, saharia2022image, chong2024catvtonconcatenationneedvirtual}, we investigate the use of LDMs to generate high-quality Virtual Try-Off images, aiming to surpass the output quality of traditional GAN-based as well as the recently proposed LDM based methods. Conventional image-to-image translation techniques typically encode the input image with a variational autoencoder and use a denoising U-Net [29] to guide the diffusion process. Additionally, many models incorporate pre-trained encoders, such as CLIP [26], to enable text-prompted transformations. However, these approaches add significant complexity, requiring large numbers of trainable parameters, which complicates and slows down training.

\begin{figure}
\centering
\setkeys{Gin}{width=0.38\linewidth}
\subfloat[Binary cloth mask ]{\includegraphics{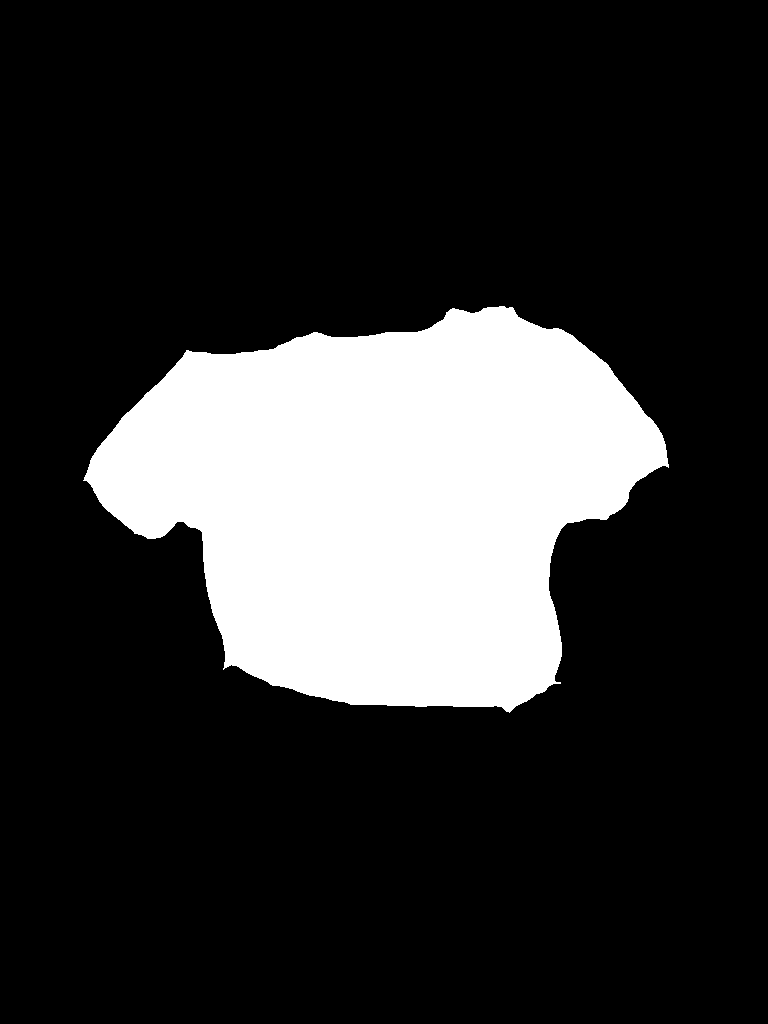}}\hfil
\subfloat[Masked cloth image ]{\includegraphics{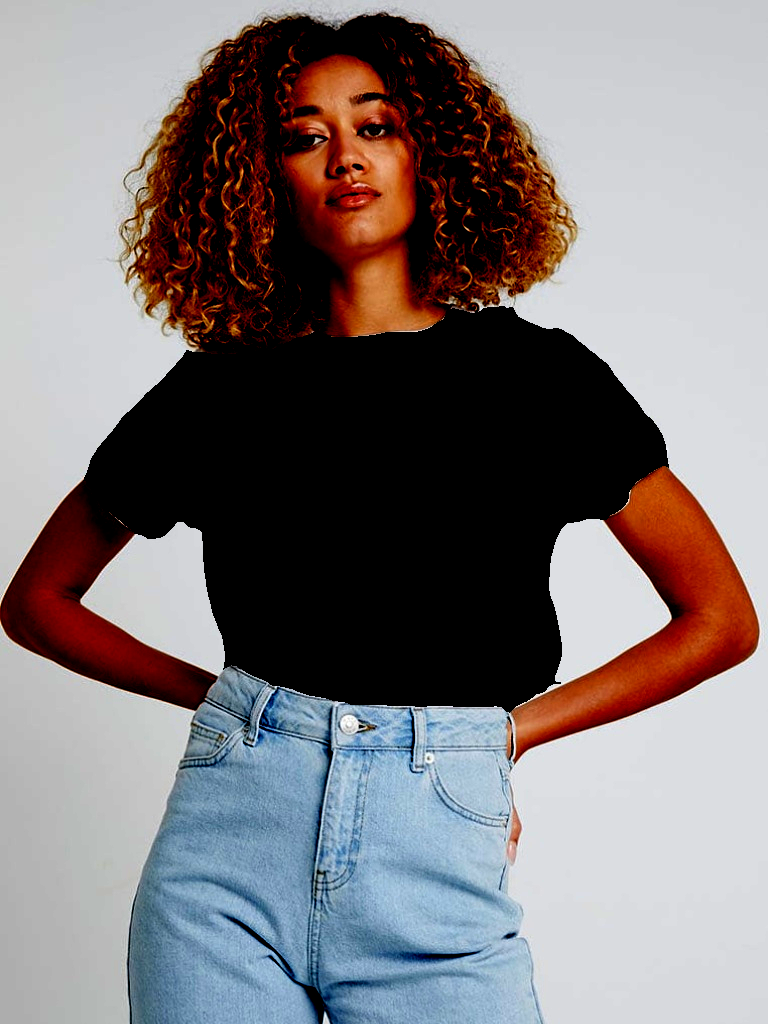}}\hfil
\caption{\centering Cloth masking}
\label{fig:mask}
\end{figure}

In this paper, we propose a streamlined single-stage model centered on a denoising U-Net. We utilize a pre-trained StableDiffusion v1.5 \cite{rombach2022high} and fine-tune our model to specifically optimize it for generating high-fidelity tiled cloth images while minimizing both network architecture complexity and training resource demands. The human-worn garment image undergoes initial processing by a variational autoencoder, yielding a latent representation that is subsequently fed into a denoising U-Net. To enhance specificity, we concatenate a latent representation of a garment mask (Figure \ref{fig:mask}) with the human-worn cloth image latents. This mask isolates the target garment, allowing the model to focus on generating a lay-down view for that specific piece. The mask additionally provides critical positional information, further improving generation quality by focusing on the desired garment’s location. Moreover, the mask enables us to generate lay-down views for both full-body and half-body input images in comparison to prior work \cite{velioglu2024tryoffdiffvirtualtryoffhighfidelitygarment} which is only capable of handling half-body images. 

To further reduce the number of trainable parameters of the network, we have removed cross-attention layers that typically facilitate text-image interaction in the transformer blocks of the U-Net \cite{chong2024catvtonconcatenationneedvirtual}. Since our model does not rely on textual input for guidance, these cross-attention layers were deemed unnecessary for the application of generating tiled cloth views. The exclusion of these layers not only simplifies the model architecture but also significantly reduces the memory and computational load, making the model more efficient and accessible for training on limited resources.

Additionally, we conducted a detailed analysis to identify which components of the U-Net are most critical for achieving high-quality tiled cloth image generation. Through experimentation, we discovered that training only the transformer blocks within the U-Net yields optimal generation quality. This selective training approach allows us to retain the generative capacity of the network while reducing the number of trainable parameters substantially. By isolating and training only the transformer blocks, we are producing realistic and structurally accurate lay-down cloth images with fewer computational requirements in contrast to prior work \cite{velioglu2024tryoffdiffvirtualtryoffhighfidelitygarment} that fine-tunes the complete U-Net.

Specifically, we make the following contributions:
\begin{itemize}
    \item We propose a Latent Diffusion Model (LDM) tailored specifically for the task of generating tiled cloth images from human-worn garment inputs (Virtual Try-Off).
    
    \item We present a simple, single-stage network that given a human-worn garment image and a corresponding mask as inputs, can produce the lay-down image of any garment while handling both half-body and full-body input images.

    \item We achieve state-of-the-art results utilizing a pre-trained StableDiffusion model and fine-tuning it to the specific task by only training specific network layers, reducing the overall trainable parameter count to 267.24M.

    \item We conduct a thorough experimental analysis, demonstrating that our proposed method consistently generates high-quality, high-fidelity tiled cloth images. Our approach outperforms existing methods, as validated on the benchmark dataset VITON-HD.

\end{itemize}

This paper is organized as follows: In Section \ref{sec:related} we review the related work. Section \ref{sec:preliminaries} discusses the foundational components of our proposal. Section \ref{sec:tiled} introduces our novel diffusion-based image lay-down model, while \ref{sec:experimental} presents our thorough experimental evaluation. Section \ref{sec:conc} concludes the paper by discussing avenues to future work in this area.

\begin{figure*}[h!]
\centering
\includegraphics[height=0.49\linewidth]{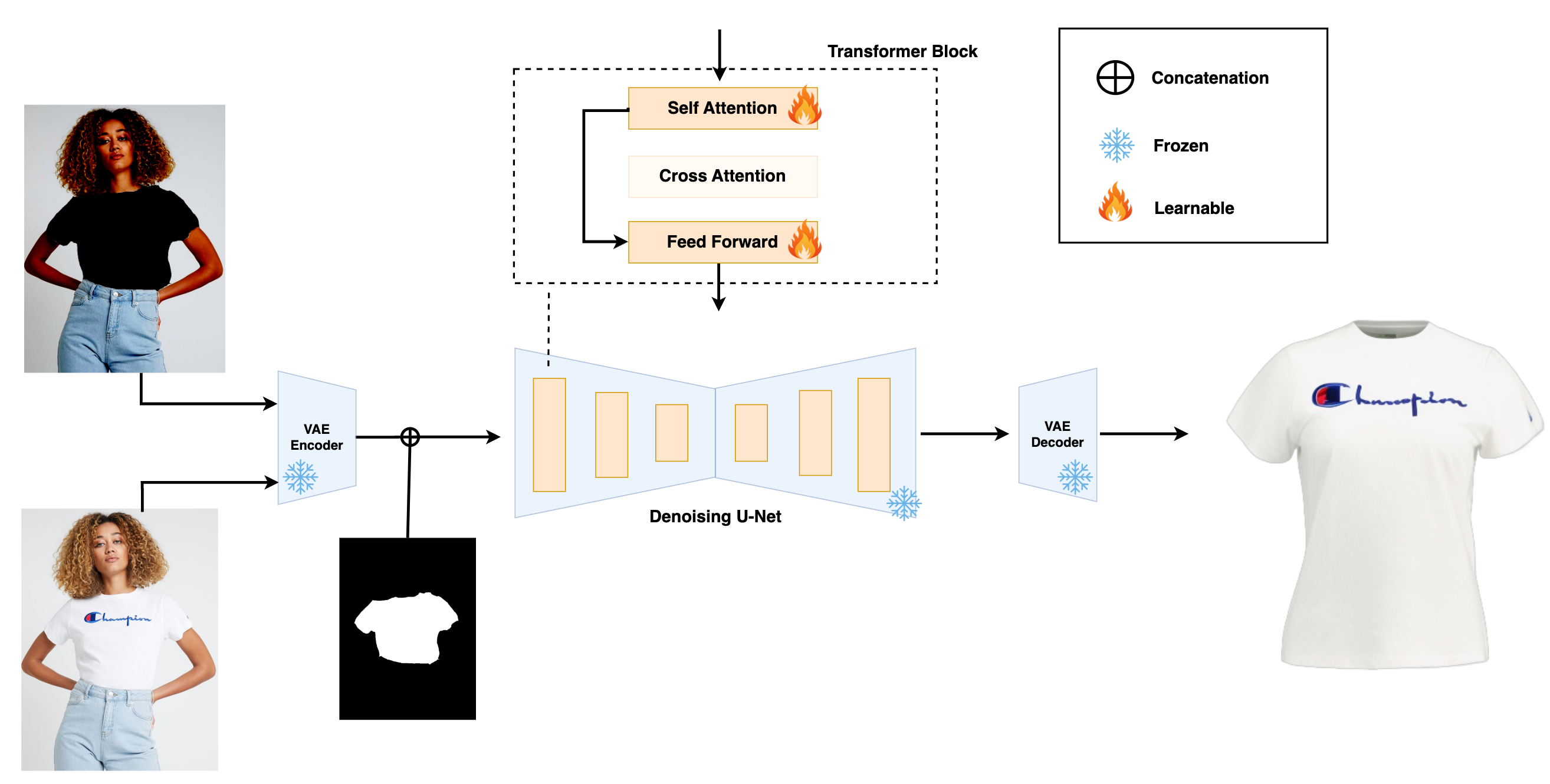}
\caption{TryOffAnyone Network Architecture}
\label{fig:overall}
\end{figure*}

\section{Related Work}
\label{sec:related}

\subsection{Image Generation}
In the realm of image generation, particularly within text-to-image tasks, diffusion models \cite{ho2020denoising, dhariwal2021diffusion, nichol2021improved, saharia2022photorealistic, rombach2022high, balaji2022ediffi, ramesh2022hierarchical, saharia2022palette} have emerged as foundational architectures due to their robust semantic priors, developed through extensive training on large-scale datasets. Efforts to adapt these models for subject-driven image generation have led to several notable approaches. Paint by Example \cite{paintbyexample} replaces text conditions with image conditions, utilizing a CLIP image encoder to prompt image generation; while it enables image-to-image generation, it often preserves the original subjects' features only approximately. IP-Adapter \cite{ipadapter} injects CLIP image features into a pre-trained latent diffusion model via an adapter, facilitating plug-and-play subject-to-image generation, though it shares similar limitations in precise feature preservation. DreamBooth \cite{dreambooth} refines diffusion models by introducing specific vocabulary to denote particular subjects, achieving consistent subject-driven text-to-image generation, but necessitates fine-tuning the model for each new subject. AnyDoor \cite{anydoor} employs DINOv2 and a ControlNet conditioned on high-frequency maps to achieve relatively accurate subject-driven image generation without extensive model adjustments. MS-Diffusion \cite{msdiffusion} leverages layout guidance to facilitate multi-subject zero-shot image personalization, offering a flexible and efficient approach to generating personalized images without prior subject-specific training. BLIP-Diffusion \cite{blipdiffusion} introduces a multimodal encoder pre-trained to provide subject representation, enabling zero-shot subject-driven generation and efficient fine-tuning for customized subjects. FastComposer \cite{fastcomposer}, a tuning-free method for multi-subject image generation, employs localized attention mechanisms to maintain subject fidelity without requiring fine-tuning. Subject-Diffusion \cite{subjectdiffusion} enables open-domain personalized text-to-image generation without test-time fine-tuning by utilizing a novel framework that combines text and image semantics.

\subsection{Image-to-image translation}

Image-to-image translation methods aim to modify specific regions of an image while preserving the surrounding context, such as the background. These approaches have been widely explored using both GAN-based and diffusion-based frameworks. GAN-based methods, such as Pix2Pix \cite{pix2pix}, leverage conditional GANs for paired image-to-image tasks, demonstrating strong performance in applications like style transfer and semantic segmentation. On the other hand, CycleGAN \cite{cyclegan} extends this framework to unpaired image-to-image translation by introducing a cycle-consistency loss, which ensures structural coherence between the input and translated images. Despite their success, GAN-based methods often struggle with issues like mode collapse and unstable training dynamics, which can limit their scalability to more complex tasks.

Recently, diffusion-based methods have emerged as a powerful alternative for image-to-image translation, offering significant improvements in fidelity and controllability. Stochastic Differential Editing (SDEdit) \cite{sdedit} edits images by inferring a latent variable at an intermediate time step and synthesizing a target image through the reverse stochastic differential equation. DiffusionCLIP \cite{diffusionclip} fine-tunes pre-trained text-to-image generative networks using local directional CLIP loss \cite{DBLP:journals/corr/abs-2103-00020} while preserving background details through an L1 reconstruction loss. This approach also employs face identity loss for human face editing tasks. Imagic \cite{imagic} optimizes pre-trained diffusion models by conditioning on inferred text features to faithfully reconstruct input images. During inference, it synthesizes target images by linearly combining the features of the source and target text, enabling fine-grained controllability. Blended Diffusion \cite{blendeddiffusion} employs user-provided masks to combine latent features of source and target images, selectively editing regions of interest while preserving the background. Palette \cite{saharia2022palette} extends diffusion models to various image-to-image tasks, such as inpainting, super-resolution, and colorization, using a denoising process to ensure high-quality reconstructions. InstructPix2Pix \cite{instructpix2pix} introduces a novel framework that enables fine-grained image editing guided by textual instructions, combining CLIP embeddings with pre-trained diffusion models to align edits with user intents.

\subsubsection{Virtual Try-Off}
The generation of images of garments laid flat, from human-worn cloth images, has garnered some attention in fashion computer vision. TileGAN \cite{zeng2020tilegan} introduces a two-stage framework for high-fidelity garment image synthesis in image-to-image translation. The first stage employs a U-Net-inspired encoder-decoder with skip connections to transfer shared features like color and shape, enhanced by a spatial transformer module to handle pose variations and spatial misalignments efficiently. Additionally, a one-hot encoded clothing category vector is integrated into the latent representation, ensuring the network generates garments of the desired type, even in complex multi-garment scenarios. The second stage refines the coarse outputs using a modified Pix2Pix framework with a dual-path attention generator. This generator incorporates a channel attention module to suppress redundant information (e.g., human body features) and a self-attention module to capture global dependencies across the image. TileGAN also uses a PatchGAN \cite{PatchGan} discriminator, which evaluates image patches to ensure fine-grained realism. By combining spatial transformations, category supervision, and attention mechanisms, TileGAN delivers detailed, photorealistic results, addressing challenges in pose complexity and garment diversity effectively.

TryOffDiff \cite{velioglu2024tryoffdiffvirtualtryoffhighfidelitygarment} is built on the Stable Diffusion (v1.4) model, a latent diffusion framework originally designed for text-to-image generation, but adapted for image-guided generation by conditioning the process on reference images. The model uses SigLIP \cite{zhai2023sigmoidlosslanguageimage}, an enhanced version of CLIP’s Vision Transformer (ViT) \cite{radford2021learningtransferablevisualmodels}, to extract detailed and domain-specific visual features, retaining the full sequence of token representations in its final layer to preserve spatial information essential for fine-grained garment reconstruction. These features are processed by an adapter module, which includes a transformer encoder, linear projection, and layer normalization, before being integrated into the denoising U-Net of Stable Diffusion via cross-attention. Keys and values for the attention mechanism are derived from the processed image features, enabling precise alignment between the input image and the generated output. To optimize performance, only the adapter modules and the denoising U-Net are fine-tuned, while the SigLIP image encoder and the VAE encoder/decoder remain frozen, maintaining the robust capabilities of the pretrained components while adapting the generative layers for garment reconstruction tasks.
\section{Preliminaries}
\label{sec:preliminaries}

\subsection{U-Net}

The U-Net architecture is a widely adopted deep learning framework designed primarily for image segmentation and has since been extended to various image-to-image translation tasks. U-Net follows an encoder-decoder structure, where the encoder extracts hierarchical features from the input, and the decoder reconstructs the output based on these features. A defining characteristic of U-Net is the use of skip connections between corresponding layers in the encoder and decoder, allowing the network to retain high-resolution spatial information while leveraging abstract semantic features. This architecture ensures that fine-grained details are preserved, which is particularly critical in tasks requiring precise localization and pixel-level accuracy, such as the Virtual Try-Off.


\subsection{Stable Diffusion}
TryOffAnyone fine-tunes Stable Diffusion \cite{rombach2022high}, a Latent Diffusion Model (LDM) trained on LAION dataset \cite{schuhmann2022laion5b}, to generate tiled cloth images from human-worn cloth images. LDMs map high-dimensional image inputs to a lower-dimensional latent space using a pre-trained Variational Autoencoder (VAE) \cite{kingma2013vae}, reducing computational cost while maintaining high-quality outputs.

An LDM comprises a denoising UNet $E_{\theta}(\cdot, t)$ and a VAE with an encoder $\epsilon$ and decoder $D$. The training minimizes:

\begin{equation}
\mathcal{L}_{\text{LDM}} := \mathbb{E}_{\epsilon(x), \epsilon \sim \mathcal{N}(0, 1), t} 
\left[\| \epsilon - \epsilon_{\theta}(z_t, t) \|_2^2\right],
\end{equation}

where \( z_t \) represents the noisy latent encoding at timestep \( t \). The forward process adds Gaussian noise \( \mathcal{N}(0, 1) \) to the encoded input, while the reverse process iteratively denoises \( z_t \) to reconstruct \( z_0 \). Finally, \( D \) decodes \( z_0 \) back into the image domain.


\section{TryOffAnyone}
\label{sec:tiled}
TryOffAnyone is a novel single stage framework designed to synthesize high-quality tiled cloth images (Figure \ref{fig:garment}a) from an input image of a dressed person (Figure \ref{fig:garment}b) and a corresponding cloth mask that covers the garment region (Figure \ref{fig:mask}b). We utilize the Segformer semantic segmentation model \cite{DBLP:journals/corr/abs-2105-15203} fine-tuned on the ATR dataset \cite{ATR} to extract accurate cloth masks. The cloth mask serves as explicit guidance for the generation process, indicating the precise garment to be reconstructed in a lay-flat configuration. This approach addresses challenges posed by complex scenes in the input image, such as the presence of multiple garments or occlusions, ensuring accurate extraction and generation of the desired clothing item.

Unlike Stable Diffusion \cite{rombach2022high}, which relies on text encoders such as CLIP \cite{DBLP:journals/corr/abs-2103-00020} to condition the generation process, our proposed architecture eliminates the need for textual descriptions by leveraging the spatial information provided by the cloth mask. This mask-based guidance significantly improves the generation process in two key aspects; it enhances image quality by providing explicit localization of the target garment, reducing ambiguities introduced by textual prompts, while it streamlines the training pipeline by removing the computational overhead associated with text encoders, thereby improving training efficiency.

The architecture of the TryOffAnyone consists of two primary modules: a pre-trained Variational Autoencoder (VAE) \cite{kingma2013vae} and a denoising U-Net \cite{ronneberger2015u}. The VAE operates as a latent space encoder-decoder, compressing the input image into a lower-dimensional representation while preserving essential garment features. The U-Net, conditioned on the dressed person's cloth and the cloth mask latent representations, performs iterative denoising to generate the target tiled cloth image.

\subsection{Network Architecture}
In Figure \ref{fig:overall}, we illustrate the proposed network architecture, highlighting our fine-tuning approach for the transformer blocks within the denoising diffusion U-Net. To identify the optimal configuration, we experimented with various training parameter setups, including fine-tuning the entire architecture, the transformer blocks, and only the attention layers within the transformer blocks.

After extensive experimentation, we selected transformer block fine-tuning as the most effective approach, yielding significantly higher performance compared to attention-only and full-architecture training. Fine-tuning only the attention layers proved insufficient for the complex task of generating virtual try-ons, as it lacked the capacity to fully capture the intricate structural and textural details required. On the other hand, while fine-tuning the full architecture achieved comparable results to transformer block fine-tuning, it required substantially more computational resources.

The transformer block fine-tuning approach strikes an ideal balance, particularly on smaller benchmark datasets such as VITON-HD. We demonstrate that focusing on transformer blocks not only achieves comparable performance to fine-tuning the full U-Net but also significantly reduces the number of trainable parameters, from 815.45 million to 267.24 million. This reduction in parameters translates to lower memory requirements, making our solution more efficient and scalable without compromising quality.

\subsection{Input Latents}

During training, the network is taking as input the human-worn cloth image $HC$ $\in \mathbb{R}^{3 \times H \times W}$ (Figure \ref{fig:garment}b) , the cloth mask $M$ $\in \mathbb{R}^{H \times W}$ (Figure \ref{fig:mask}a) , the human-worn cloth masked $HM$ $\in \mathbb{R}^{3 \times H \times W}$ (Figure \ref{fig:mask}b) and the desired output cloth image in lay-down view $C$ $\in \mathbb{R}^{3 \times H \times W}$(Figure \ref{fig:garment}a).

The inputs $HC$ and $HM$ are then encoded into the latent space using a shared VAE encoder :

\[
X_{HC} = VAE(HC)
\]
\[
X_{HM} = VAE(HM)
\]

where $X_{HC}, X_{HM} \in \mathbb{R}^{4 \times \frac{H}{8} \times \frac{W}{8}}$. Then, the latents of the human worn cloth image $X_{HC}$ and the human-worn cloth masked $X_{HM}$ are concatenated along the spatial dimensions to form:

\[
X = [X_{HC}, X_{HM}],
\]
where $X \in \mathbb{R}^{4 \times \frac{H}{8} \times \frac{W}{8}}$.

The mask $M$ is concatenated with an all-zero mask of the same size:

\[
X_M = [M, zeros(M)],
\]
where $X_M \in \mathbb{R}^{ \frac{H}{8} \times \frac{W}{8}}$.

Then, we encode the garment image cloth $C$ into the latent space using the variational autoencoder $VAE$ 

\[
X_{C} = VAE(C)
\]

We proceed by concatenating the latents of the garment image cloth $X_C$ with the human-worn cloth image $X_{HC}$

\[
X_{CM} = [X_{C}, X_{HC}],
\]
where $X_{CM} \in \mathbb{R}^{4 \times \frac{H}{8} \times \frac{W}{8}}$, and apply random noise on the concatenated $X_{CM}$ latents.

\[
N = Noise(X_{CM})
\]
Finally, we concatenate the latents $X$, $X_M$, and the noise $N$ into $I$
\[
I = [N, X_M, X],
\]
where $I \in \mathbb{R}^{9 \times \frac{H}{8} \times \frac{W}{8}}$ and provide them to the U-Net as input to obtain a denoised representation.

\section{Experimental Evaluation}
In this section, we present the results of a thorough experimental evaluation of our proposals. To ensure reproducibility and provide a comprehensive evaluation, we meticulously detail our experimental setup. For assessing the quality of generated outputs, we employ several quantitative evaluation metrics, as well as qualitative comparisons on two benchmark datasets. We conduct an extensive ablation study to rigorously evaluate the effectiveness of our proposed methods across various configurations. This includes a detailed analysis of the impact of seed numbers on the quality and consistency of the generated lay-down images, offering valuable insights into the factors influencing model performance.

\label{sec:experimental}

\subsection{Datasets}

Our study utilizes two datasets; the publicly available VITON-HD \cite{vton-hd}, and a dataset provided by FULL BEAUTY BRANDS (FBB). These datasets serve as robust benchmarks for the virtual try-off task due to their comprehensive coverage of diverse garment types and use cases. The VITON-HD dataset comprises of 13,679 high-resolution (1024 x 768) image pairs featuring frontal half-body models and their corresponding upper-body garments. Originally curated for the Virtual Try-On (VTON) task, the dataset is also well-suited for our study as it provides the necessary image pairs ($C$, $HC$), where $C$ is the garment in lay-down image view (Figure \ref{fig:garment}a), and $HC$ is the human-worn cloth image (Figure \ref{fig:garment}b). We follow the same cleaning preprocessing employed in TryOffDiff \cite{velioglu2024tryoffdiffvirtualtryoffhighfidelitygarment}, to eliminate duplicates and resolving train/test leakage. After preprocessing, the cleaned dataset comprises 11,552 image pairs for training and 1,990 image pairs for testing.


The FBB dataset consists of a total of 24,568 images, covering a diverse range of garment categories. Specifically, the dataset includes 12,771 images of upper-body garments (e.g., T-shirts, shirts, blouses), 7,158 images of bottoms (e.g., pants, skirts), and 4,639 images of dresses. To create a test set for evaluation purposes, 10\% of the images from each category are reserved, resulting in a comprehensive dataset split that supports rigorous testing while maintaining a substantial training set for model development. Additionally, the FBB dataset features a greater diversity of images, including both full-body and half-body shots, a wide variety of poses, that increase the complexity of the try-off task, and varying image resolutions. 

Across all datasets, we utilize the original image pairs depicting individuals in various poses and their corresponding in-shop garments. However, instead of relying on traditional methods such as DensePose \cite{DensePose} for clothing mask generation, we employ the SegFormer \cite{DBLP:journals/corr/abs-2105-15203} model to create the clothing masks. This approach ensured more precise and detailed segmentation of the garments, contributing to improved downstream performance. This consistent masking methodology was applied to both the VITON-HD and FBB datasets, contributing to enhanced model performance.

\subsection{Metrics}
\label{sec:metrics}

For paired virtual try-off settings with ground truth in the test datasets, we utilize five widely adopted metrics to assess the similarity between synthesized images and real images; Structural Similarity Index (SSIM) \cite{SSIM}, Fréchet Inception Distance (FID) \cite{Seitzer2020FID}, Kernel Inception Distance (KID) \cite{bińkowski2021demystifyingmmdgans}, Learned Perceptual Image Patch Similarity (LPIPS) \cite{LPIPS}, and DISTS \cite{DISTS}. Our test dataset encompasses a broad range of garment categories, including T-shirts, Shirts, Blouses, Pants, Skirts, and Dresses. To ensure a comprehensive evaluation, we calculate the average SSIM, FID, and DISTS across these categories. Among these metrics, DISTS, which evaluates image similarity by incorporating both perceptual and structural differences, is regarded as the most significant for assessing output quality. A comprehensive analysis conducted by TryOffDiff \cite{velioglu2024tryoffdiffvirtualtryoffhighfidelitygarment} demonstrated that DISTS is particularly well-suited for the virtual try-off task, as it effectively captures visual fidelity and fine-grained details in synthesized images. Consequently, we prioritize DISTS in evaluating the performance of our models.

\subsection{Implementation Details}
We trained our models using the inpainting variant of the pre-trained StableDiffusion v1.5 model \cite{rombach2022high}. To ensure a comprehensive evaluation and maintain a fair comparison, we independently trained separate models for the VITON-HD \cite{vton-hd} and FBB datasets, and their respective test sets were utilized for quantitative and qualitative evaluations. To train our model, we employed the AdamW optimizer \cite{loshchilov2017decoupled}, setting a batch size of 4, a fixed learning rate of $1 \times 10^{-5}$, and the DDIM \cite{DBLP:journals/corr/abs-2010-02502} noise scheduler. The models were trained for 20 epochs at a resolution of 512×384. The experiments were conducted on a cluster of 4 NVIDIA A10G GPUs, taking approximately 8 hours to train each model.

\subsection{Baseline Approaches}

TileGan \cite{zeng2020tilegan} evaluates its performance on a custom dataset similar to VITON-HD, but with several additional cloth categories. To ensure a fair comparison, we restrict our evaluation of TileGan to t-shirts, shirts, and blouses that are included in the VITON-HD test set. Unfortunately, qualitative comparisons with TileGan were not possible due to the unavailability of its code and model. However, from the qualitative results reported in their work, it is evident that our method demonstrates superior performance. While TileGan successfully captures the general shape of garments, it struggles with intricate details, such as texture fidelity and complex patterns.

For TryOffDiff \cite{velioglu2024tryoffdiffvirtualtryoffhighfidelitygarment}, we utilized their publicly available quantitative metrics on VITON-HD to benchmark our results against theirs. To enable qualitative comparisons, we leveraged their online demo \cite{tryoff} to generate images. This allowed us to evaluate TryOffDiff visual output and compare its qualitative performance with our approach. By integrating both their metrics and demo outputs, we ensured a thorough and fair assessment of TryOffDiff capabilities in the virtual try-off task.

\subsection{Quantitative Comparison}
In Table \ref{table:quantitative}, we present a comprehensive quantitative analysis of the performance of our proposed TryOffAnyone method compared to state-of-the-art approaches, including TryOffDiff \cite{velioglu2024tryoffdiffvirtualtryoffhighfidelitygarment} and TileGan \cite{zeng2020tilegan}, on the VITON-HD dataset \cite{vton-hd}. We do not include results for TryOffDiff or TileGan on the FBB dataset. TryOffDiff cannot be applied directly, as it includes both full-body and half-body images, along with garment categories (i.e. outerwear, pants, dresses, skirts) which fall outside the capabilities of TryOffDiff. Furthermore, TileGan has not publicly released its code or pretrained models.

Our analysis highlights that TryOffAnyone consistently surpasses TryOffDiff across several key metrics, reinforcing its robustness and efficacy for virtual try-off applications. Specifically, our method achieves superior performance on the DISTS metric, which is deemed as the most appropriate measure for evaluating results in the VTOFF task \cite{velioglu2024tryoffdiffvirtualtryoffhighfidelitygarment}. The enhanced DISTS score demonstrates the superior capability of TryOffAnyone in preserving garment fidelity and visual coherence.

Additionally, our proposed method demonstrates signficant improvements on the LPIPS and KID metrics, further underscoring its effectiveness in capturing fine-grained details and achieving high perceptual similarity. Notably, our approach achieves comparable performance to TryOffDiff on the FID metric, suggesting that both methods produce similarly realistic outputs when assessed by this widely used generative model evaluation standard.

However, it is worth noting that TryOffDiff achieves a significantly higher score on the SSIM metric, indicating stronger structural similarity with ground-truth images. This discrepancy requires further investigation to understand the underlying reasons and explore potential avenues for improvement. Factors such as the sensitivity of SSIM to luminance and contrast variations, or the potential overemphasis on structural regularity in TryOffDiff, may contribute to this result. 

When compared to TileGan, our proposed method exhibits marked advancements in performance on both the SSIM and FID metrics, highlighting its ability to outperform a method designed primarily for tile-based generative tasks. This improvement demonstrates the adaptability and robustness of TryOffAnyone in handling complex garment structures and diverse image compositions, which are critical for achieving high-quality virtual try-on results.

\begin{table}
\caption{\label{table:quantitative} Quantitative comparison against TryOffDiff and TileGan on VITON-HD dataset}
\begin{center}
\resizebox{\columnwidth}{!}{%
\begin{tabular}{|c|c|c|c|c|c|}
\hline
 & SSIM ↑ & FID ↓ & KID ↓ & LPIPS ↓ & DISTS ↓\\ \hline
TryOffDiff (s=42) & \textbf{79.5} & \textbf{25.1} & 8.9 & 32.4 & 23.0 \\ \hline
TileGan & 70.96 & 39.8 & N/A & N/A & N/A \\ \hline
TryOffAnyone (s=36) & 71.99 & 25.3 & \textbf{2.01} & \textbf{17.22} & \textbf{21.01} \\ \hline
\end{tabular}%
}
\end{center}
\end{table}

\subsection{Qualitative Comparison}
\begin{figure*}
\centering
\includegraphics[width=0.94\linewidth]{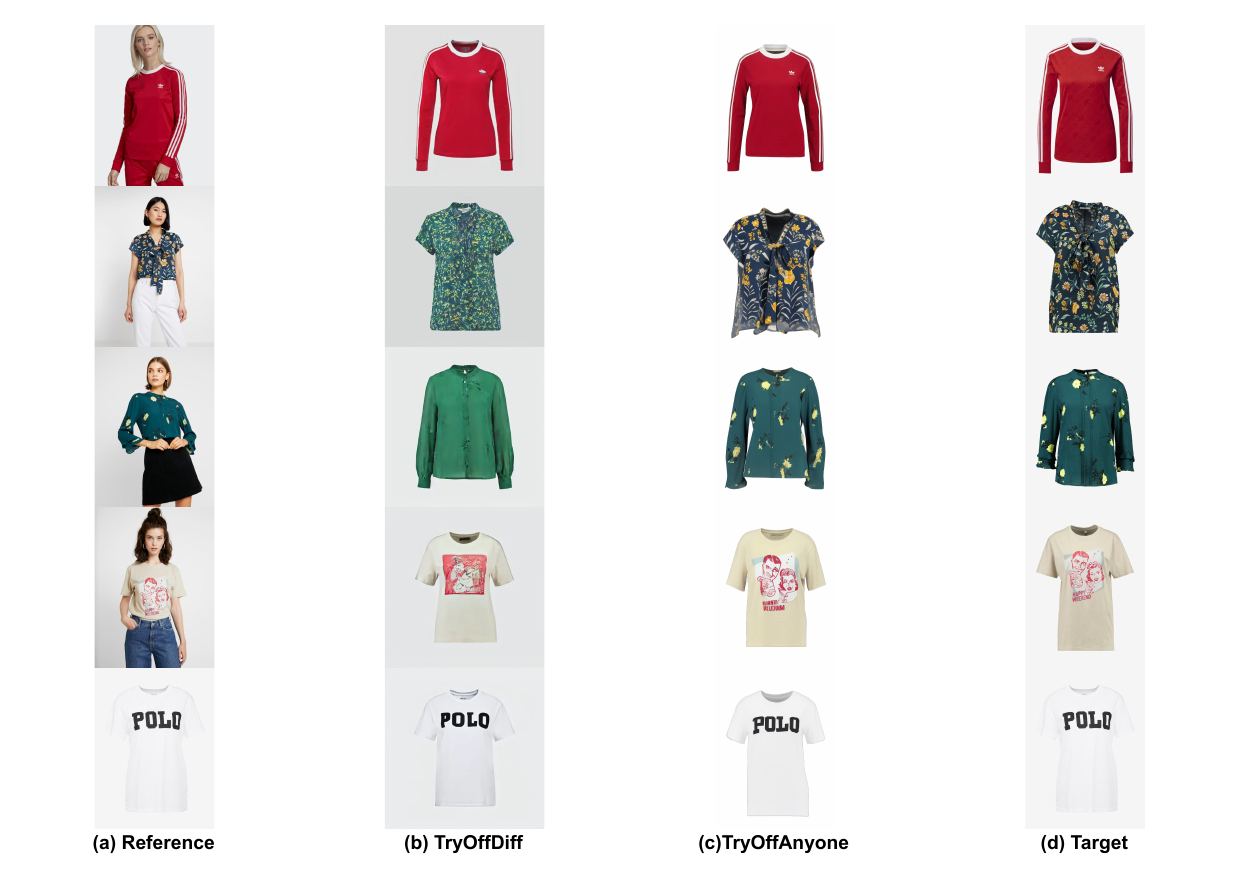}
\caption{Qualitative comparison against TryOffDiff \cite{velioglu2024tryoffdiffvirtualtryoffhighfidelitygarment} on VITON-HD \cite{vton-hd}}
\label{fig:qualitative_comp}
\end{figure*}

In Figure \ref{fig:qualitative_comp}, we present a comparative analysis of our generated lay-down images alongside those produced by TryOffDiff \cite{velioglu2024tryoffdiffvirtualtryoffhighfidelitygarment} on the VITON-HD dataset. Our method demonstrates a clear and consistent advantage in terms of both texture fidelity and pattern accuracy, setting a new benchmark for the virtual try-off task.

While TryOffDiff achieves comparable performance when reconstructing garments with logos, it often struggles to accurately represent garments with intricate patterns. Specifically, TryOffDiff tends to capture only the general structure and color of these garments, leaving finer details of the pattern underrepresented. In contrast, our approach delivers precise and detailed renderings of garments across all instances, including those with highly complex and irregular patterns, making a more appropriate solution for the virtual try-off task.

\begin{figure*}
\centering
\includegraphics[width=0.98\linewidth]{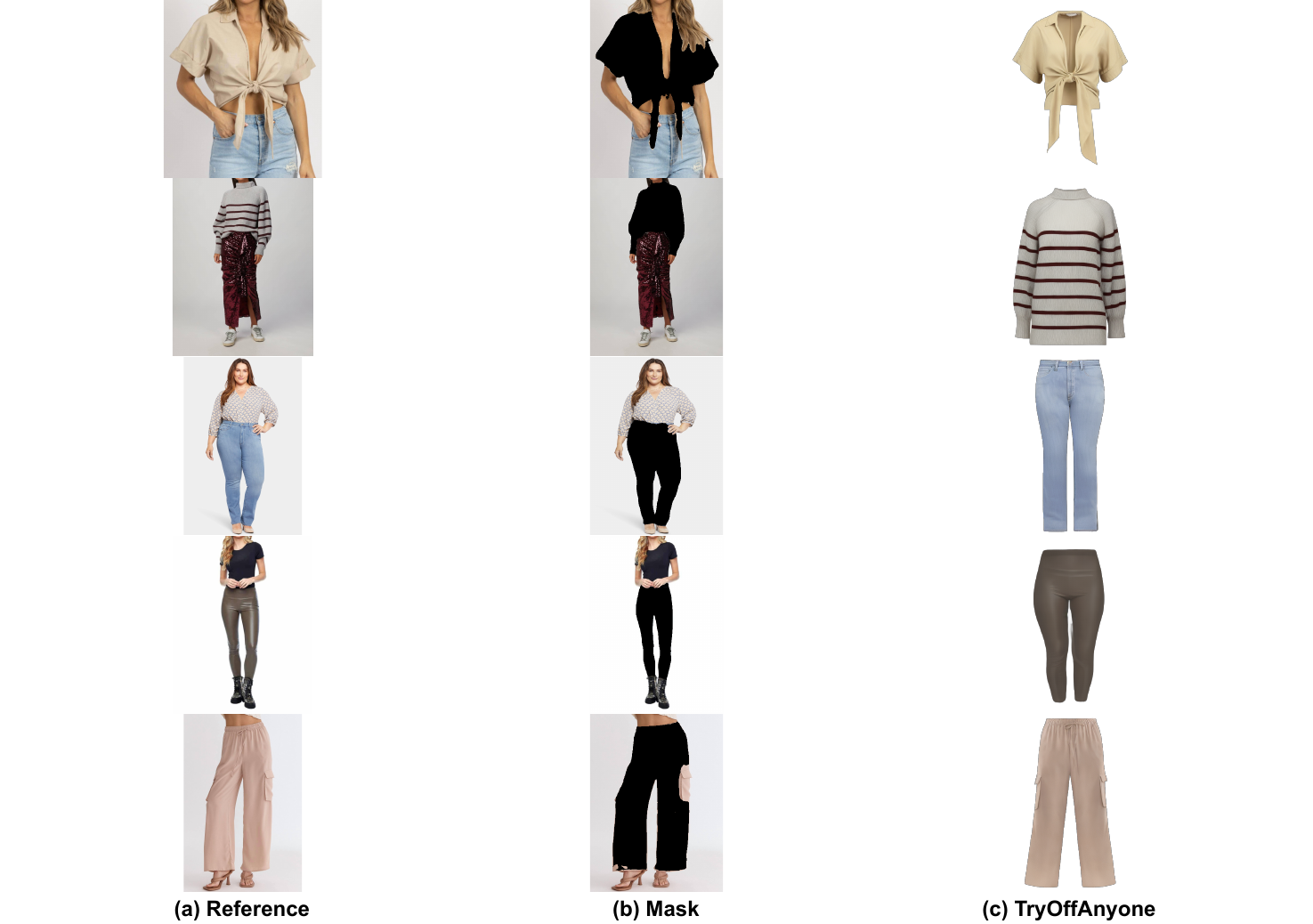}
\caption{Qualitative results on FBB dataset}
\label{fig:qualitative_comp_dress}
\end{figure*}

In Figure \ref{fig:qualitative_comp_dress}, we illustrate the qualitative performance of our proposed TryOffAnyone model on the FBB dataset. Unlike TryOffDiff, which lacks the ability to guide the diffusion process to target specific garments for laydown generation, our model is fully capable of handling this task. This limitation renders TryOffDiff unsuitable for datasets like FBB dataset, which feature full-body human-worn garment images.

Our TryOffAnyone model demonstrates exceptional performance on FBB dataset, achieving highly accurate laydown reconstructions across multiple dimensions, including garment structure, color, texture, and intricate pattern details. Similar to its performance on VITON-HD, our model excels at capturing even the most subtle features of garments, ensuring that the generated laydowns are both realistic and true to the original garment characteristics.

\subsection{Ablation Studies}
In this section, we present a comprehensive ablation study to validate and justify the design choices of our proposed model. The study examines multiple facets, including variations in network training parameter configurations, a comparative evaluation of text-guided laydown generation against our mask-guided architecture, and an analysis of the impact of different seed numbers used to guide the initial inference noise on the final tiled cloth outputs.

First, we explore the effects of different network training parameter setups to understand their influence on the model’s performance. This analysis highlights the optimal configurations that balance computational efficiency and reconstruction quality, ensuring the best outcomes for garment laydown generation.

Second, we evaluate a text-guided laydown generation model, where textual prompts replace mask inputs, against our mask-guided architecture. While the text-guided approach provides a flexible alternative, our results demonstrate that the mask-guided method achieves superior performance.

Lastly, we conduct a detailed analysis of the effects of varying seed numbers, which guide the initial noise during inference. This evaluation reveals how different seeds impact the quality and consistency of the final tiled cloth output. The results show that certain seeds yield more detailed and visually consistent outputs, while others introduce artifacts or inconsistencies.

\subsubsection{Training Parameter Setup}

In this experiment, we evaluate the impact of different model configurations on the performance of the virtual try-off task. We investigate three scenarios: (a) fine-tuning the full denoising U-Net architecture, (b) fine-tuning only the transformer blocks within the U-Net, and (c) fine-tuning the self-attention layers within the transformer blocks. The results of these setups, evaluated on the VITON-HD dataset using five metrics outlined in Section \ref{sec:metrics}, are summarized in Table \ref{table:settings}.

The results reveal that fine-tuning only the self-attention layers (setup c) yields significantly lower performance compared to the other two configurations. This underperformance can be attributed to the complexity of the virtual try-off task, which involves intricate image translation challenges. Factors such as occlusions caused by human poses and variations in lighting conditions demand a model setup capable of capturing fine-grained details of garments.

In contrast, setups (a) and (b) exhibit comparable performance. Notably, training only the transformer blocks reduces the number of trainable parameters by a factor of four, making it a more computationally efficient choice without sacrificing accuracy. Consequently, we adopt setup (b) for our experimental evaluations. However, it is worth noting that fine-tuning the full U-Net architecture (setup a) may offer performance gains when applied to larger datasets than VITON-HD, suggesting potential avenues for future exploration in scaling this approach.

\begin{table}
\caption{\label{table:settings} Performance on different training settings}
\begin{center}
\resizebox{\columnwidth}{!}{%
\begin{tabular}{|c|c|c|c|c|c|}
\hline
Setup (Parameters) & SSIM ↑ & FID ↓ & KID ↓ & LPIPS ↓ & DISTS ↓\\ \hline
Full U-Net (815.45M) & 71.22 & \textbf{25.11} & \textbf{2.0} & 18.51 & \textbf{20.85} \\ \hline
Transformer (267.24M) & \textbf{71.99} & 25.3 & 2.01 & \textbf{17.22} & 21.01 \\ \hline
Attention (49.57M)& 68.38 & 87.8 & 5.1 & 21.66 & 22.94 \\ \hline
\end{tabular}%
}
\end{center}
\end{table}



\subsubsection{Mask-Guided vs Text-Guided Try-Off}
To validate the effectiveness of our mask-guided tiled cloth image generation approach, we developed a comparative text-guided model, replacing the mask inputs with textual prompts such as “Generate a lay-down image for the Top garment” or “Generate a lay-down image for the Bottom garment”. We utilize the InstructPix2Pix model \cite{instructpix2pix} pre-trained on Stable Diffusion v1.5 \cite{rombach2022high}. We fine-tune the whole U-Net architecture, including the cross-attention layers, on the VITON-HD dataset. This setup allowed us to assess the impact of mask guidance versus text-based guidance on the laydown image generation process. The results of this comparison are presented in Table \ref{table:text_vs_mask}.

Our evaluation highlights the superiority of the proposed mask-guided model, which consistently outperforms the text-guided alternative across all metrics. While both models achieve similar scores on structural similarity (SSIM) and perceptual similarity (LPIPS), which primarily measure how well the structure of the garment is preserved, the mask-guided approach demonstrates clear advantages in reconstructing fine details. Specifically, the positional information embedded in the cloth mask enables the diffusion process to better capture nuanced details such as color accuracy, texture fidelity, and intricate patterns.

These results underscore the advantages of our mask-guided architecture for virtual try-off tasks. By providing explicit spatial guidance, the mask not only enhances detail recovery but also improves overall image quality, making it a more reliable and effective method for generating accurate and realistic garment lay-down images.

\begin{table}
\caption{\label{table:text_vs_mask} Text vs Mask guidance}
\begin{center}
\resizebox{\columnwidth}{!}{%
\begin{tabular}{|c|c|c|c|c|c|}
\hline
Guidance & SSIM ↑ & FID ↓ & KID ↓ & LPIPS ↓ & DISTS ↓\\ \hline
Text Guidance & 70.8 & 62.3 & 6.33 & 22.8 & 22.56 \\ \hline
Mask Guidance & \textbf{71.99} & \textbf{25.3} & \textbf{2.01} & \textbf{17.22} & \textbf{21.01} \\ \hline
\end{tabular}%
}
\end{center}
\end{table}

\subsubsection{Seed Number Variation}
\begin{figure*}
\centering
\includegraphics[width=0.94\linewidth]{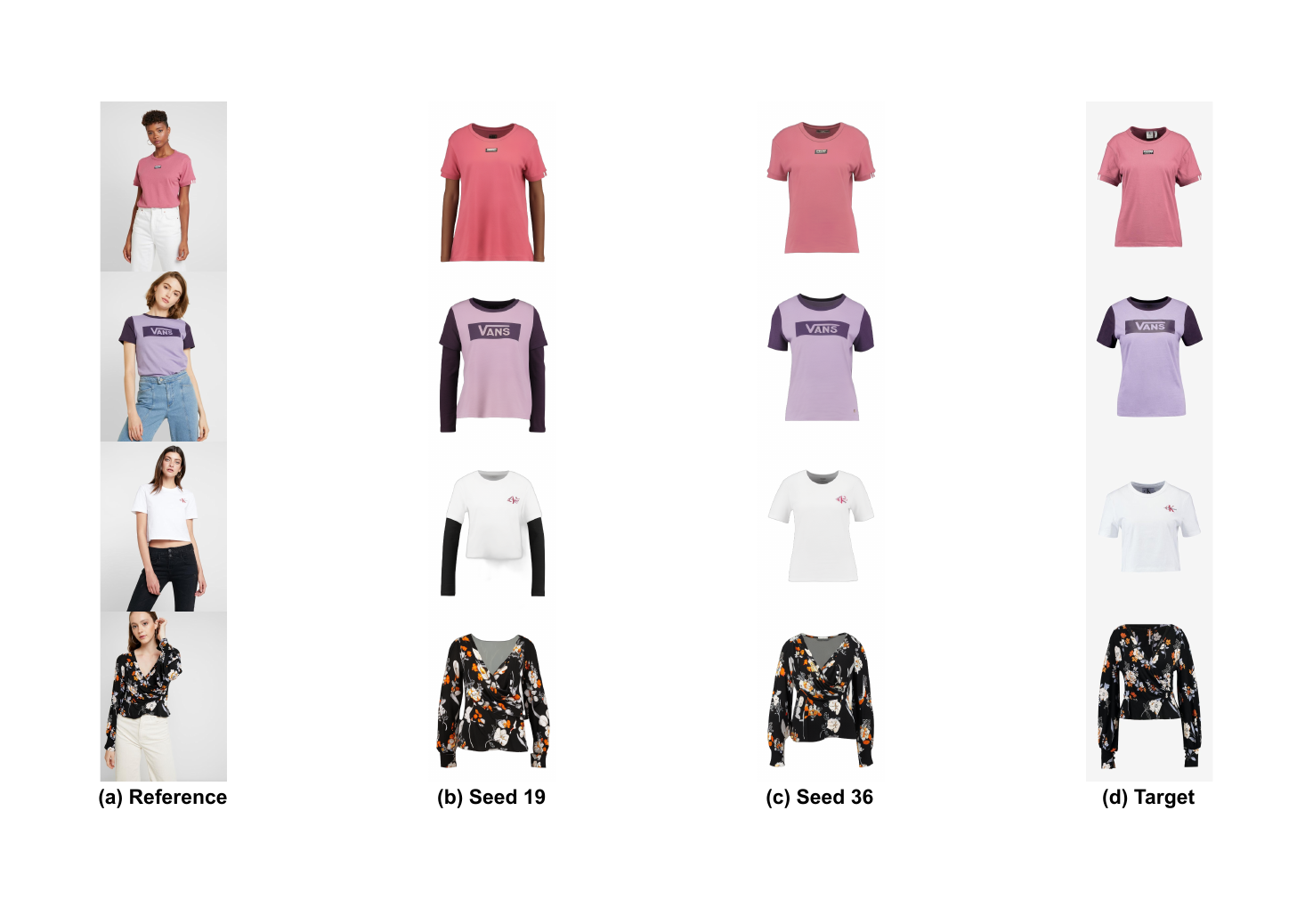}
\caption{TryOffAnyone Seed Number Comparison on VITON-HD}
\label{fig:seed_comparison}
\end{figure*}
In this section, we evaluate the impact of different random seeds on the model’s performance. Recently the seed selection performance on text-to-image tasks was evaluated by Xu et al. \cite{xu2024good}, showing that certain seed numbers produce significantly better outcomes. In our experimentation, we notice that the seed selection on the virtual try-off task plays a significant role in the final image quality. To this end, we conduct a large-scale experiment, where we compare the effect of 100 different seed numbers. Interestingly, we observe significant differences in the model’s results, both quantitatively and qualitatively, depending on the chosen seed. These variations highlight the sensitivity of the inference process to initialization conditions, which can influence the model’s ability to generalize effectively. 

In Tables \ref{table:best_seed} and \ref{table:worst_seed}, we present a quantitative comparison of different seed numbers. Table \ref{table:best_seed} showcases the three seeds that achieved the best performance, while Table \ref{table:worst_seed} highlights the three seeds with the poorest performance. The evaluation focuses on the DISTS metric. We observe that the three best seeds achieve significant performance benefits, in comparison with seeds 52, 36, and 94 which achieve the best DISTS score performance. To validate the quantitative performance difference of alternative seeds, in Figure \ref{fig:seed_comparison}, we showcase visual examples of different outputs based on the seed number. As we observe, the best-performing seed (seed=36) accurately depicts the lay-down view of the clothes in all instances. However, the worst-performing seed (seed=19) tends to depict the laydowns with limbs or converts short sleeves to long sleeves t-shirts. In the last instance, we observe that when the input garment is long-sleeved, the worst-performing seed performs extremely well, similar to seed 36.

\begin{table}
\caption{\label{table:best_seed} Best performing seed numbers}
\begin{center}
\resizebox{\columnwidth}{!}{%
\begin{tabular}{|c|c|c|c|c|c|}
\hline
Seed \# & SSIM ↑ & FID ↓ & KID ↓ & LPIPS ↓ & DISTS ↓\\ \hline
\small{Seed 52} & 71.46 & 30.5 & \textbf{2.0} & \textbf{17.11} & \textbf{20.53} \\ \hline
\small{Seed 36} & \textbf{71.99} & \textbf{25.3} & 2.01 & 17.22 & 21.01 \\ \hline
\small{Seed 94} & 71.5 & 32.4 & 3.3 & 17.4 & 21.09 \\ \hline
\end{tabular}%
}
\end{center}
\end{table}

\begin{table}
\caption{\label{table:worst_seed} Worst performing seed numbers}
\begin{center}
\resizebox{\columnwidth}{!}{%
\begin{tabular}{|c|c|c|c|c|c|}
\hline
Seed \# & SSIM ↑ & FID ↓ & KID ↓ & LPIPS ↓ & DISTS ↓\\ \hline
\small{Seed 79} & 68.69 & 49.14 & 4.26 & 20.28 & 23.26 \\ \hline
\small{Seed 7} & 65.80 & 60.13 & \textbf{5.28} & 22.33 & 23.60 \\ \hline
\small{Seed 19} & \textbf{65.21} & \textbf{101.08} & 4.75 & \textbf{23.66} & \textbf{24.10} \\ \hline
\end{tabular}%
}
\end{center}
\end{table}

\section{Conclusions}
\label{sec:conc}

In this work, we presented a streamlined and efficient approach to generating high-fidelity tiled garment images, addressing a critical need in fashion-focused computer vision applications. By leveraging the power of Latent Diffusion Models (LDMs) and a fine-tuned StableDiffusion network, we achieved state-of-the-art results in producing realistic and structurally accurate garment images from human-worn photographs. Our innovative use of garment masks enabled the model to isolate and focus on specific clothing items, significantly enhancing the quality of the generated outputs for both full-body and half-body inputs. Experimental evaluations on VITON-HD dataset validated the superiority of our method over existing approaches, showcasing its potential for scalable deployment in e-commerce and virtual try-on systems. 

This work opens numerous avenues for further study. Improving the texture quality of the lay-down images using a refiner network is a natural extension of the present work. Additionally, further experimentation and exploration are required to identify the reasons behind different seed numbers perform better in the task of try-off as well as in other image generation applications.

\bibliographystyle{ACM-Reference-Format}
\bibliography{sample-sigconf}

\end{document}